\documentclass[letterpaper, 10 pt, conference]{ieeeconf}  

\IEEEoverridecommandlockouts                              

\usepackage{graphics} 
\usepackage{epstopdf}
\usepackage{diagbox}
\usepackage{xcolor}
\definecolor{my_stage1}{RGB}{180,199,231}
\definecolor{my_stage2}{RGB}{255,217,102}
\definecolor{train_shape}{RGB}{143,170,220}
\definecolor{test_shape}{RGB}{244,177,131}

\usepackage{epsfig} 
\usepackage{times} 
\usepackage{multirow}
\usepackage{amsmath} 
\usepackage{makecell}
\usepackage{amsfonts}            
\usepackage{bm} 
\usepackage{CJK} 
\usepackage{geometry}
\newcommand{\figref}[1]{Fig.~\ref{#1}}

\usepackage{subfigure}
\usepackage{cite}
\usepackage{gensymb}
\usepackage[colorlinks,
linkcolor=black,
anchorcolor=black,
citecolor=black]
{hyperref}

\title{\LARGE \bf
Multi-Modal Fusion in Contact-Rich Precise Tasks via Hierarchical Policy Learning
}
\author{Piaopiao Jin$^{1}$, Yinjie Lin$^{2}$, Yanchao Tan$^{3}$, Tiefeng Li$^{1}$ and Wei Yang$^{1}$%
\thanks{$^{1}$ Piaopiao Jin, Tiefeng Li and Wei Yang are with Faculty of Center for X-Mechanics, Department of Engineering Mechanics, Zhejiang University, Hangzhou 310027, China.
    {\tt\small piaopiaojin@zju.edu.cn, litiefeng@zju.edu.cn, yangw@zju.edu.cn}}%
\thanks{$^{2}$ Yinjie Lin is with the State Key Laboratory of Fluid Power \& Mechatronic Systems, Zhejiang University, Hangzhou 310027, China.
    {\tt\small yjlin@zju.edu.cn@zju.edu.cn}}%
\thanks{$^{3}$ Yanchao Tan is with the College of Computer Science, Zhejiang University, Hangzhou 310027, China.
    {\tt\small yctan@zju.edu.cn.}}%
}

\begin{document}
\maketitle
\thispagestyle{empty}
\pagestyle{empty}

\begin{abstract}
Combined visual and force feedback play an essential role in contact-rich robotic manipulation tasks. Current methods focus on developing the feedback control around a single modality while underrating the synergy of the sensors. Fusing different sensor modalities is necessary but remains challenging. A key challenge is to achieve an effective multi-modal and generalized control scheme to novel objects with precision.
This paper proposes a practical multi-modal sensor fusion mechanism using hierarchical policy learning.
To begin with, we use a self-supervised encoder that extracts multi-view visual features and a hybrid motion/force controller that regulates force behaviors.
Next, the multi-modality fusion is simplified by hierarchical integration of the vision, force, and proprioceptive data in the reinforcement learning (RL) algorithm.
Moreover, with hierarchical policy learning, the control scheme can exploit the visual feedback limits and explore the contribution of individual modality in precise tasks. 
Experiments indicate that robots with the control scheme could assemble objects with \textit{0.25mm} clearance in simulation. The system could be generalized to widely varied initial configurations and new shapes. Experiments validate that the simulated system can be robustly transferred to reality without fine-tuning.
\end{abstract}
\section{INTRODUCTION}
The automation of robots is highly valued for its increasingly important roles in our daily lives, especially in contact-rich tasks, like grasping, pick and place, and assembly. To achieve the goal, the controllers that are able to leverage vision or force feedback are usually devised.
The vision modality provides a global, semantic and geometric description of the scene\cite{DBLP:journals/corr/abs-1803-09956}\cite{andrychowicz2020learning}\cite{DBLP:journals/corr/abs-1809-08564}, while its feedback is limited under occlusion, usually in an unstructured manipulation environment. 
The force feedback, on the other hand, provides local contact information of the robot-environment interaction, and is sensitive to slight contact variations and collisions \cite{10.1115/1.3149634}\cite{7743375}\cite{luo2019reinforcement}. The force control is only valid with pre-defined regions and suffers from wide-range adaptation. 
Due to the limitations in both modalities, multi-modal fusion on feedbacks of different characteristics is an important and challenging issue to be solved and leads to the development of a controller that is robust to unforeseen conditions.

As a case study, we adopt the precise assembly task. 
To achieve the task, the robot needs both vision and force feedback.
There are mainly two challenges in achieving effective multi-modality fusion in the assembly task.

\textit{Challenge 1.} \textit{How to robustly process multi-modality information in the assembly task?} 
The commonly used modalities in the manipulation tasks include vision, force, and proprioception. 
These modalities differ in sample frequencies, value ranges, and dimensions. Therefore, the different modalities need to be specifically treated. There are mainly two approaches to processing the multiple modalities. 
The first is to decompose the task into two subtasks: the vision-based pose detection task and the force-based controller design. Gao \textit{et al}. extracted a key point representation shared among the category-level objects \cite{DBLP:journals/corr/abs-2102-06279}. The force-based controller takes the detection result as the target position command. The simple decomposition, however, ignores the intertwined proper of the modalities. 
The second method utilizes machine learning (ML) to fuse the different sensors. 
Reinforcement learning (RL) \cite{sutton2018reinforcement}, a specific kind of ML technique, allows learning the multi-modality fusion sequentially. Lee \textit{et al}. \cite{lee2020making} built a multi-modal control scheme in the assembly task. 
The fusion of the multi-modal information enhanced the performance when compared with the single-modality control scheme.
However, doubts remain for the uncertainties in the vision and force feedback: including the occlusion and the inaccurate force feedback in the simulated environment.

\textit{Challenge 2.} \textit{How to effectively fuse multi-modal information with RL?}
RL endows robots the promise to fuse multi-modalities and to accommodate variations in environmental configurations.
Studies solving multi-modal manipulation tasks with RL have been presented \cite{lee2020making}\cite{luo2021robust}. In these methods, a multi-modal representation is learned for the assembly task. 
Due to the differences in the modalities, it is important to select certain modality encoders.
For example, the raw image is usually encoded with a convolutional neural network (CNN) \cite{lecun1995convolutional}; and the high-frequency force feedback is encoded with a sequence model like a recurrent neural network (RNN) \cite{vaswani2017attention}. The resultant multi-modal observation space is usually high-dimensional and challenging to train.
Besides, understanding the the individual role of each modality would help us better fuse the modalities in precise manipulation tasks.
How to make the best use of the vision in the contact-rich tasks and how to evaluate the performance increase with the force incorporation remains unclear.

\begin{figure*}[thpb]
    \centering
    \includegraphics[scale=0.40]{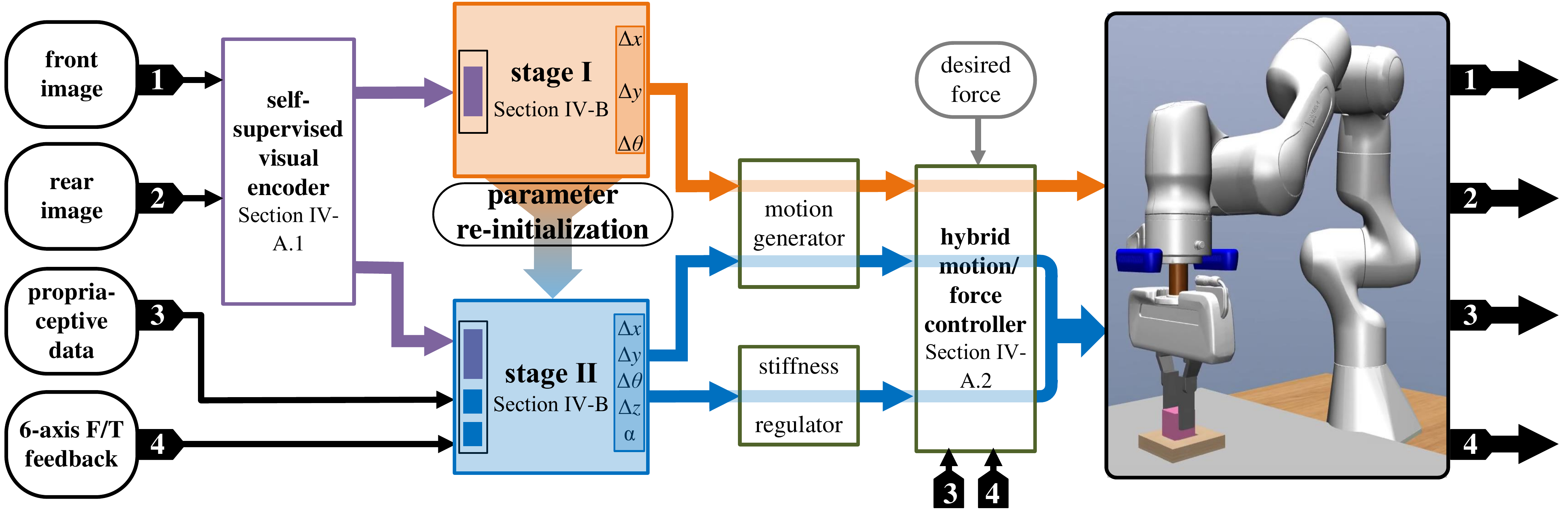}
    \caption{
    Overview of the multi-modal fusion with hierarchical policy learning in contact-rich tasks.
        }
    \label{fig1}
 \end{figure*}
To address the two challenges mentioned above, we propose a hierarchical policy learning approach of the multi-modal control scheme in contact-rich tasks (shown in \figref{fig1}).

Specifically, we separately process the vision, force, and proprioceptive data with a self-supervised image encoder and a hybrid motion/force controller. Two difficulties remain for direct pose detection.
Firstly, the manipulation precision requirement usually exceeds the detection limits, and the manipulation scene is usually partially occluded. Therefore, getting a precise pose of the target place is unrealistic. To effectively extract visual features, a self-supervised visual encoder is adopted to extract spatial relationships between the grasped object and the hole. (Sec. \ref{subsec:visual encoder})
Secondly, the force feedback is usually inaccurate in simulated environments due to the reduced contact model. Moreover, the motion and the force response of the system in precise tasks vary in different directions. To address the two issues, a robust hybrid motion/force controller is proposed. (Sec. \ref{subsec:force controller})

To enhance the effectiveness of multi-modal control scheme learning, 
we devise a hierarchical policy learning structure. 
The multi-modality fusion is divided into two stages. In the first stage, a visuomotor policy is developed with the visual features.
After making the best use of the visual feedback, force and proprioceptive data are integrated into the RL algorithm. The policy outputs are also more informative with a stiffness regulator in the force controller.
Hierarchical policy learning simplifies the construction of modality encoders and promotes the interpretability of the contribution of a single modality. (Sec. \ref{subsec:par2})

The contribution of this work could be summarized as follows.
\begin{itemize}
    \item{\textbf{Practical Multi-modality Fusion Scheme}}
    We propose a hierarchical multi-modal fusion framework where a contact-rich manipulation policy can be learned.  
    
    \item{\textbf{Effective Model Design}}
    We separately propose a self-supervised image encoder and a hybrid motion/force controller for robust multi-modal information procession.
    A hierarchical policy learning procedure for effective multi-modal fusion is devised.
    
    \item{\textbf{Experiments Verifications}}
    Extensive experiments are conducted to verify the proposed method. 
    The control scheme could be generalized to largely varied initial positions and novel objects. 
    The multi-modal control scheme could be transferred to real scenarios robustly. 
\end{itemize}
\section{RELATED WORK}
\subsection{Multi-modal Sensing in Manipulation}
Heterogeneous sensor modalities hold the promise of providing more informative feedback for solving manipulation tasks than uni-modal approaches. Specifically, visual and tactile data \cite{li2019connecting} have been utilized in object cluttering \cite{zhang2020visual}, grasp assessment\cite{van2016stable} \cite{Calandra_2018}, and pose detection \cite{9561222}. 
Beside the visual and tactile data fusion, the compound concerning visual and force/torque feedbacks is also under study. Gao \textit{et al}. \cite{DBLP:journals/corr/abs-2102-06279} used a visual-based keypoint detector and a force controller to perform category-level manipulations.
Along with the successful procession of the visual and force feedbacks, the fusion of the two modalities is actively studied. Lee \textit{et al}. utilized a multi-modal differentiable particle filter for state estimation \cite{lee2020multimodal}. They even proposed an effective multi-modal representation \cite{lee2021detect} when input modalities are corrupted. 
However, the hierarchical relationship between visual and force has rarely been studied in contact-rich precise manipulation tasks.

\subsection{RL-based Manipulation}
Recent advances in RL demonstrate successful applications in contact-rich manipulation tasks, such as grasping, non-prehensile pushing, and assembly.
RL endows robots the promise to accommodate variations in environmental configurations.
Dong \textit{et al}. \cite{dong2021icra} developed a tactile-based RL algorithm for insertion tasks. Oikawa used a non-diagonal stiffness matrix for precise assembly \cite{9361338}.
\begin{figure}[thpb]
    \centering
    \includegraphics[scale=0.31]{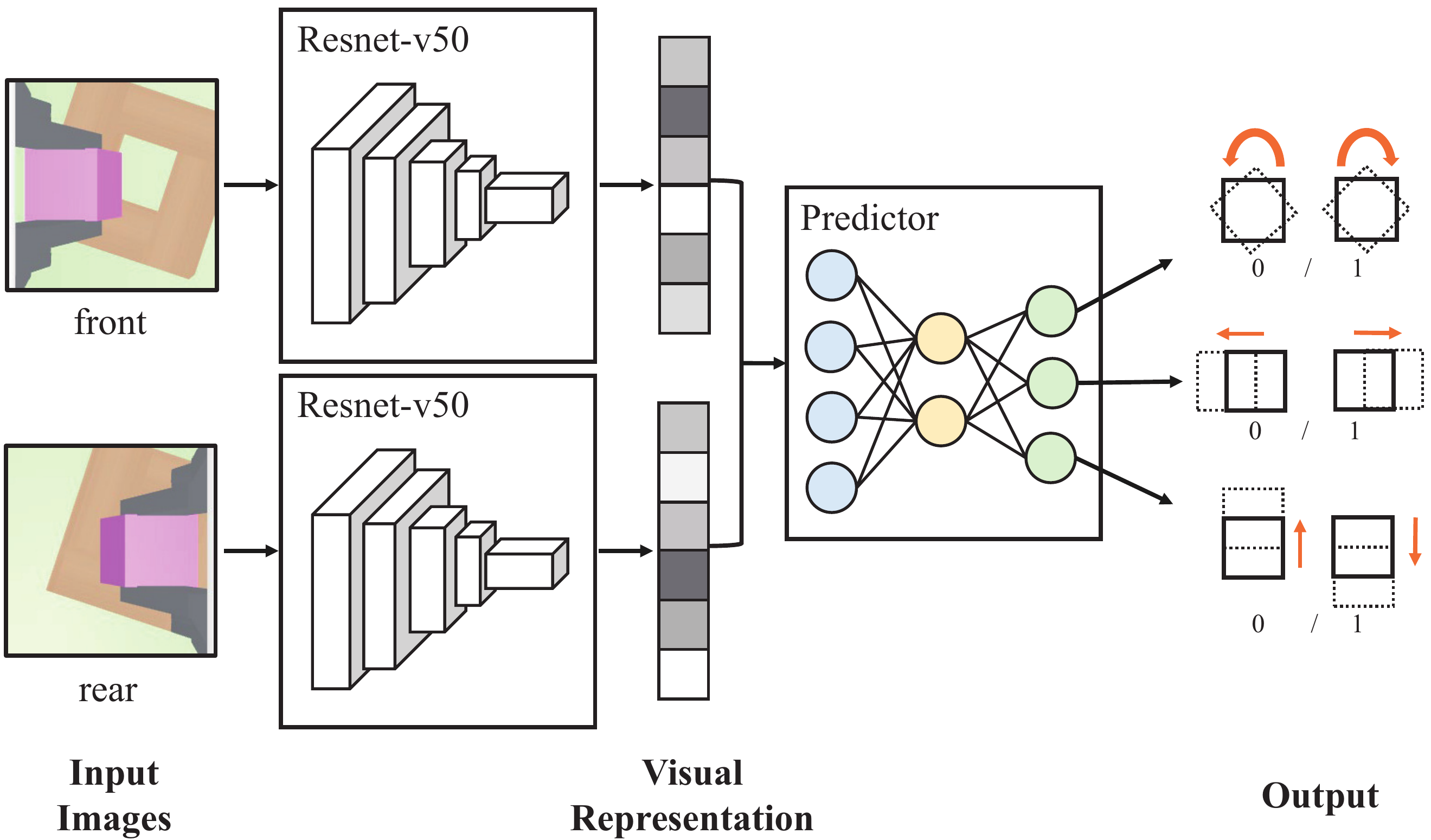}
    \caption{The neural network architecture of the self-supervised visual encoder.
    }
    \label{fig4}
\end{figure}

However, only a few works are learning multi-modal manipulation policies in RL. Narita \textit{et al}. decomposed the multi-modal manipulation into several primitive skills \cite{narita2021policy}. A resultant multi-modal controller is formalized by the single-modality feedback controller. The work most similar to ours is \cite{lee2020making}. Lee \textit{et al}. learned a representation model that combines vision, haptics, and proprioceptive data. The state representation is validated in peg-in-hole insertion tasks. The multi-modal feature is, nevertheless, complicated and requires a large amount of fine-tuning. Moreover, the controller shows weak generalization ability when transferred from simulation to reality.
In the present work, a hierarchical policy learning procedure is adopted for multi-modality fusion.
The method simplifies the construction of the modality encoders and promotes the interpretability of the single modality.
Our resultant control scheme could assemble with 0.25mm clearance, which is one magnitude smaller than the result in \cite{lee2020making}.

\section{PROBLEM STATEMENT AND METHOD OVERVIEW}
The objective of our algorithm is to achieve effective multi-modal fusion in the assembly task through hierarchical policy learning. The method is comprised of two parts: a robust multisensory procession and a hierarchical fusion mechanism. We first learn a self-supervised visual encoder (Sec. \ref{subsec:visual encoder}) and develop a hybrid motion/force controller (Sec. \ref{subsec:force controller}).
A multi-modal policy is then learned hierarchically through reinforcement learning (RL).

We model the manipulation task as a finite-horizon, discounted Markov Decision Process (MDP) 
$\bm{\mathcal{M}}$, which has a state space $\bm{\mathcal{S}}$,
an action space $\bm{\mathcal{A}}$, the state transition dynamics $\bm{\mathcal{T}}$:$\bm{\mathcal{S}}\times\bm{\mathcal{A}}\rightarrow\bm{\mathcal{S}}$, 
a reward function $\bm{r}:\bm{\mathcal{S}}\times\bm{\mathcal{A}}\rightarrow\bm{\mathbb{R}}$,
a horizon $\bm{\mathcal{T}}$, and a discount factor $\bm{\gamma} \in \bm{(0,1]}$. To
determine the optimal stochastic policy $\bm{\pi}:\bm{\mathcal{S}}\rightarrow\bm{\mathbb{P}}(\bm{\mathcal{A}})$,we
maximize the expected value of discounted reward as shown in \eqref{rl}.
\begin{equation} \label{rl}
\bm{J}(\bm{\pi})=\mathbb{\bm{E}}_{\bm{\pi}}{\sum^{T-1}_{t=0}{\bm{\gamma}^{t}\bm{r}(\bm{s_t},\bm{a_t})}}
\end{equation}

In the hierarchical policy learning (Sec. \ref{subsec:hierarchial learning}), the state, action, and reward functions vary during the two-stage training procedure. 

\section{METHOD}
\label{sec:section4}
This section comprises two parts: the procession of the multi-modality feedback and the effective multi-modal fusion using hierarchical policy learning.
\figref{fig1} illustrates the control framework.
\subsection{Multi-modal Procession}
\label{subsec:part1}
\subsubsection{Self-supervised Visual Encoder}
\label{subsec:visual encoder}

\begin{figure}[thpb]
    \centering
    \includegraphics[scale=0.13]{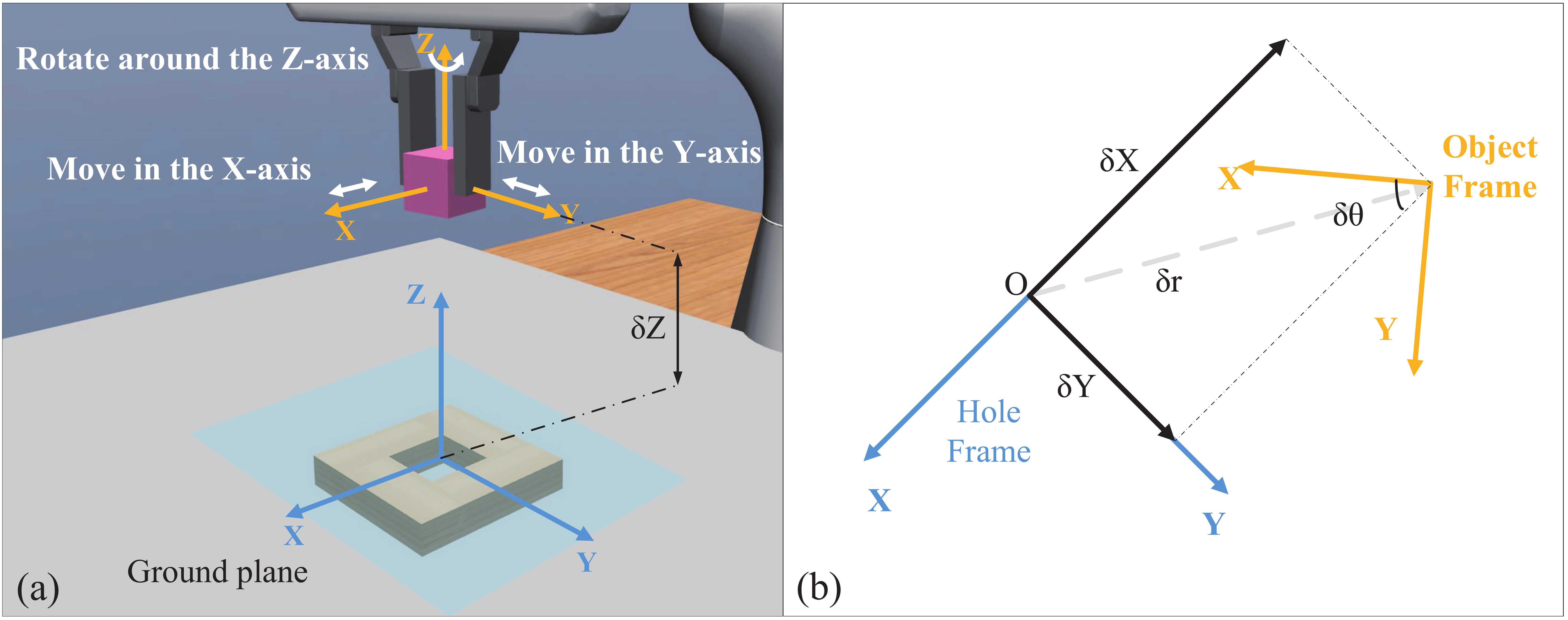}
    \caption{(a) Schematics of the hole frame and the object frame. (b) Projections of two frames on the ground plane illustrate their transformation with four parameters, $\bm{x}$, $\bm{y}$, $\bm{z}$ and $\bm{\theta}$
    }
    \label{detail_scene}
\end{figure}
The accurate pose of the hole is difficult to allocate under occlusion (the grasped object occludes the hole), especially with the lack of depth information. Therefore, a self-supervised algorithm that extracts the temporal relationship of the grasped object and the hole is proposed. The self-supervised algorithm is trained without human inference. The images and the labels are all collected in simulation automatically.

The self-supervised neural network predicts three Booleans indicating
the position and orientation relationship between the grasped object and the target hole. 
To align the object and the hole, the gripper
needs to move in the $\bm{x}-\bm{y}$ plane and to rotate about the $\bm{z}$-axis (\figref{detail_scene}). 
The two frames share the same $\bm{z}$-axis, hence the frame transformation can be represented by four parameters, $\bm{\delta}\bm{x}$, $\bm{\delta}\bm{y}$, $\bm{\delta}\bm{z}$, $\bm{\delta}\bm{\theta}$ denoting the displacements along $\bm{x}$-axis, $\bm{y}$-axis, $\bm{z}$-axis and the rotation about $\bm{z}$-axis (\figref{detail_scene}). However, the frame transformation along $\bm{z}$-axis is difficult to acquire from the two top color cameras. Therefore, the visual encoder only predicts the other three parameters. Instead of regressing to the true values of the parameters, 
the self-supervised learning algorithm predicts three Booleans, representing the positive or negative values of $\bm{\delta}$$\bm{x}$, $\bm{\delta}$$\bm{y}$, $\bm{\delta}\bm{\theta}$.
\figref{fig4} depicts the network architecture. The two collected images are first cropped into the size of 224 $\times$ 224,  and they are
separately processed by a pre-trained Resnet-v50 backbone network. The output of two networks is concatenated into a 128-d vector. The joint visual features are fed to a three-layer MLP to predict the relative displacement of the object and the hole.

\subsubsection{ARMRC-based Hybrid Motion/Force Control Design}
\label{subsec:force controller}
In RL-based contact-rich manipulation tasks, a rigid position controller is commonly used for its simplicity. However, the position controller cannot deal with the force perturbations in precise tasks. Therefore,  a hybrid motion/force controller is designed not only for motion command execution but also for force behavior modulation. Moreover, the stiffness of the system is learned in accommodation to different stages of the task.

In joint space, the system dynamics is modeled as following:
\begin{equation} \label{joint space}
    \bm{M}(\bm{q}) \ddot{\bm{q}}+\bm{C}(\bm{q}, \dot{\bm{q}}) \dot{\bm{q}}+\bm{G}(\bm{q}) + \bm{d}=\bm{\tau}+\bm{J}^{T}(\bm{q}) \bm{F}_{ext}
  \end{equation}
where $\bm{q}$ denotes the joint angle, $\dot{\bm{q}}$ the joint velocity, and $\ddot{\bm{q}}$ the joint acceleration; the symbol $\bm{M}(\bm{q})$ stands for the inertia matrix, $\bm{C}(\bm{q},\dot{\bm{q}})$ the Coriolis/centrifugal matrix and $\bm{G}(\bm{q})$ the vector of gravitational torques; the notation $\bm{d}$ represents the lumped uncertainties that include unmodeled nonlinearities and external disturbance, $\bm{\tau}$ the control input torque, $\bm{J}^{T}(\bm{q})$ the Jacobian matrix and $\bm{F}_{ext}$ the external force that exerted at the end-effector.

To design the ARMRC-based hybrid motion/force control, the operational space dynamics of end-effector is required, and it can be derived based on \eqref{joint space} as the following,
\begin{equation} \label{end-effector}
    \setlength{\abovedisplayskip}{3pt}
    \setlength{\belowdisplayskip}{3pt}
    \bm{\Lambda}(\bm{x})\ddot{\bm{x}}+\bm{\mu}(\bm{x},\dot{\bm{x}})\dot{\bm{x}}+\bm{F}_{g}+\bm{D}=\bm{F}_{\tau}+\bm{F}_{ext}
\end{equation}
where $\bm{x}$ represents the position in operational space, $\dot{\bm{x}}$ the velocity and $\ddot{\bm{x}}$ the acceleration; the symbol $\bm{\Lambda}$ denotes inertial matrix of end-effector, which can be written as $\bm{\Lambda} = \bm{J}^{\dagger T}\bm{M}\bm{J}^{\dagger}$;\footnote{Matrix $\bm{J}^{\dagger}$ is the pseudo inverse of Jacobian matrix $\bm{J}$, which satisfying the properties $\bm{J}\bm{J}^{\dagger}=\bm{I}$, $\bm{J}^{\dagger T}\bm{J}^{T}=\bm{I}$.} $\bm{\mu}(\bm{x},\dot{\bm{x}})$ the Coriolis/centrifugal matrix of end-effector in operational space, with $\bm{\mu}(\bm{x},\dot{\bm{x}})=\bm{J}^{\dagger T}(\bm{C}-\bm{M}\bm{J}^{\dagger}\dot{\bm{J}})\bm{J}^{\dagger}$; the notations $\bm{F}_{g}=\bm{J}^{\dagger T}\bm{G}$, $\bm{D}=\bm{J}^{\dagger T}\bm{d}$ and $\bm{F}_{\tau}=\bm{J}^{\dagger T}\bm{\tau}$ are the forces reflected on the end-effector.

Define the motion control subspace and force control subspace as $\bm{x}_p^T$ and $\bm{x}_f^T$ (i.e., $\bm{x}=[\bm{x}_p^T,~\bm{x}_f^T]^T$). Accordingly, the desired motion trajectory and the desired force trajectory can be written as $\bm{x}_d=[\bm{x}_{pd}^T,~\bm{x}_{fd}^T]^T$, $\bm{F}_d=[\bm{0},~\bm{f}_d^T]^T$, respectively. In order to enable the end-effector to accurately track the motion trajectory $\bm{x}_{pd}$ in motion control subspace as well as the force trajectory $\bm{f}_d$ in the constrained subspace, the desired close loop dynamics can be chosen as the following,
\begin{equation} \label{desired dynamics}
\bm{\Lambda}_d \ddot{\tilde{\bm{{x}}}}+\bm{D}_d \dot{\tilde{\bm{{x}}}}+\bm{K}_d \tilde{\bm{{x}}} =-\bm{K}_f \tilde{\bm{f}} 
\end{equation}
where $\tilde{\bm{{x}}}=[\bm{x}_p^T,~\bm{x}_f^T]^T - [\bm{x}_{pd}^T,~\bm{x}_{fd}^T]^T$ and $\tilde{\bm{f}}=\bm{F}_{ext} - [\bm{0},~\bm{f}_d^T]^T$; $\bm{\Lambda}_d,~\bm{D}_d,~\bm{K}_d,~\bm{K}_f$ are suitable matrices, which are positive definite and usually set as diagonal matrices.

With the derived end-effector dynamics \eqref{end-effector} and the desired end-effector dynamics \eqref{desired dynamics}, the objective of the rest of the subsection can be seen as determining a control input $\bm{\tau}$ so that the equation \eqref{end-effector} achieve the desired end-effector dynamics \eqref{desired dynamics}.

\begin{figure}
    \centering
    \includegraphics[scale=0.27]{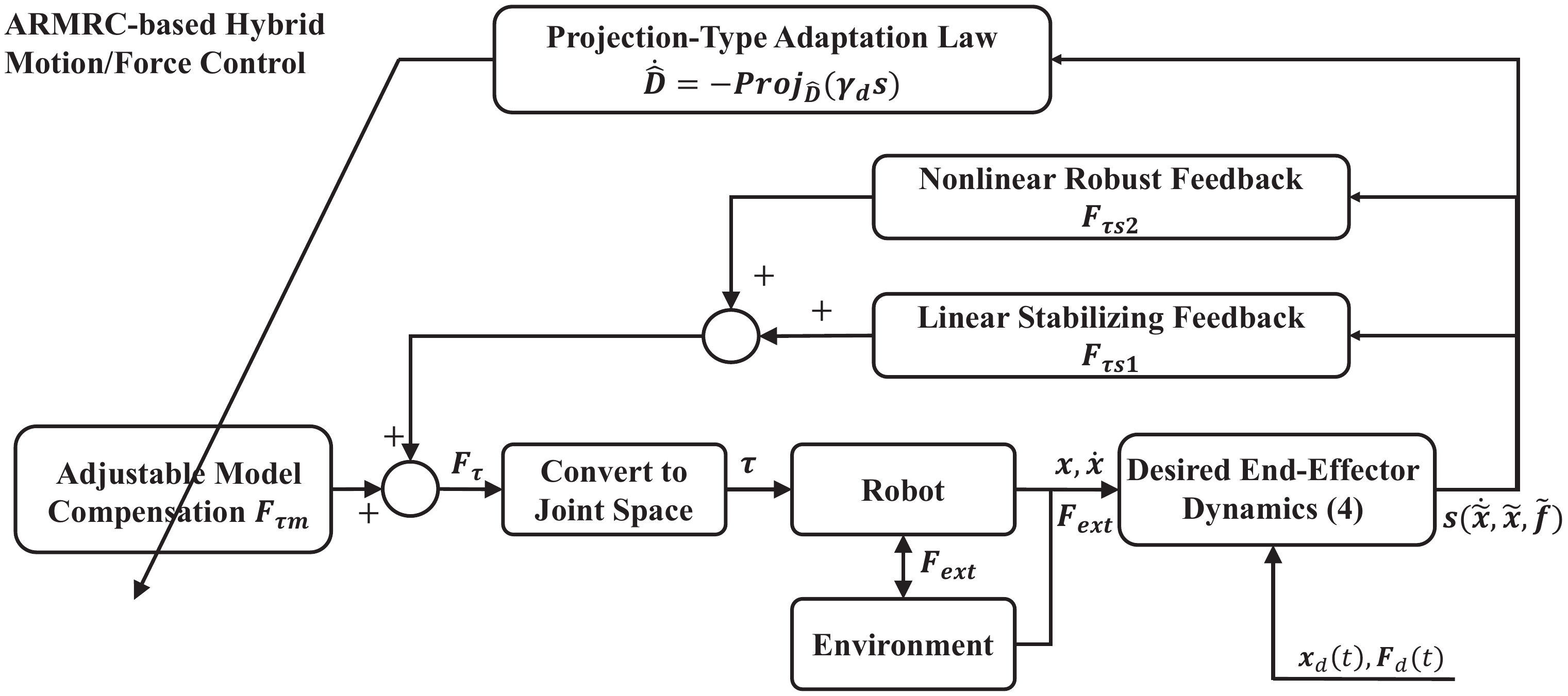}
    \caption{The structure of ARMRC-based hybrid motion/force control approach.}
    \label{fig3}
\end{figure}

As shown in the \figref{fig3}, the ARMRC law is designed to achieve the desired end-effector dynamics, which contains three terms: the linear stabilizing feedback term $\bm{F}_{\tau s1}$,  the adjustable model compensation term $\bm{F}_{\tau m}$ and the nonlinear robust feedback term $\bm{F}_{\tau s2}$. 
Specifically, we first define the switching function as 
\begin{equation} \label{s}
    \bm{s}=\dot{\tilde{\bm{{x}}}}+\bm{F}_1 \tilde{\bm{{x}}}+\bm{F}_2 \bm{z}
\end{equation}
where $\bm{F}_1$ and $\bm{F}_2$ are any positive definite diagonal matrices; $\bm{z}$ is an n-dimensional state vector and it is defined as
\begin{figure}[thpb]
    \centering
    \includegraphics[scale=0.5]{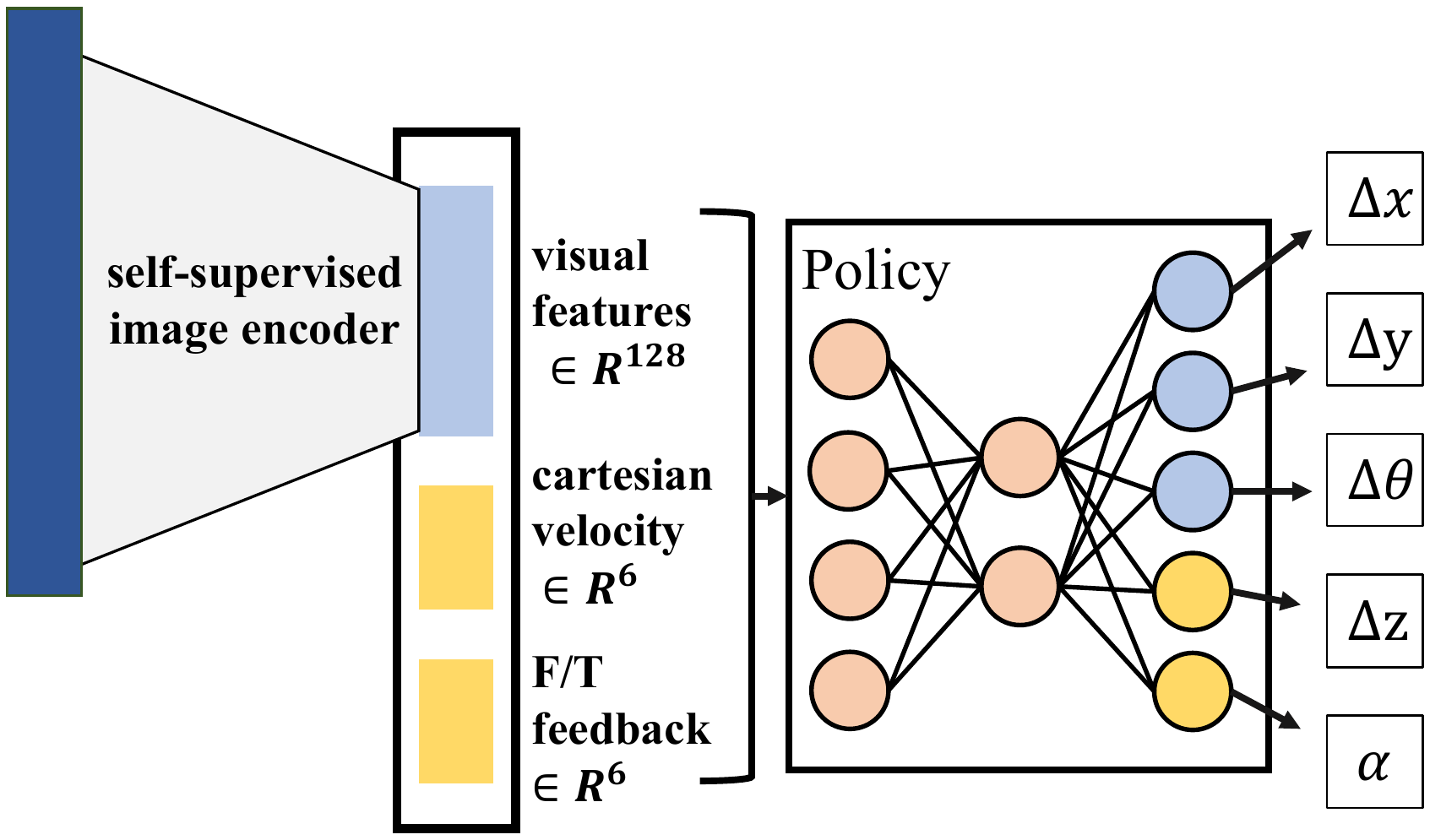}
    \caption{The schematics of the hierarchical policy learning procedure. During the first stage, the observation is 128-dimensional consisting of the visual features (\textcolor{my_stage1}{blue}). The action space is three-dimensional consisting of the desired displacement along the $\bm{x}$, $\bm{y}$ axes and rotation about the $\bm{z}$-axis (\textcolor{my_stage1}{blue}). During the second stage, the observation space is expanded with the incorporation of force and proprioceptive feedback (\textcolor{my_stage2}{orange}). The action space also includes two extra components (\textcolor{my_stage2}{orange}). One is the desired displacement along the $\bm{z}$-axis and the other is the stiffness factor along the insertion direction.}
    \label{policy}
\end{figure}
\begin{equation} \label{z}
    \begin{aligned}
    &\dot{\bm{z}}+\bm{A}\bm{z}=\bm{K}_{pz}\tilde{\bm{{x}}} + \bm{K}_{vz} \dot{\tilde{\bm{{x}}}} + \bm{K}_{fz}\tilde{\bm{f}}\\
    &\bm{K}_{v z}=\bm{F}_{2}^{-1}\left(\bm{\Lambda}_{d}^{-1} \bm{D}_{d}-\bm{F}_{1}-\bm{A}\right) \\
    &\bm{K}_{p z}=\bm{F}_{2}^{-1}\left(\bm{\Lambda}_{d}^{-1} \bm{K}_{d}-\bm{A} \bm{F}_{1}\right) \\
    &\bm{K}_{f z}=\bm{F}_{2}^{-1} \bm{\Lambda}_{d}^{-1} \bm{K}_{f}
    \end{aligned}
\end{equation}

Where $\bm{A}$ is chosen as a diagonal matrix and is positive definite.
With equations \eqref{s} and \eqref{z}, the sliding surface can be formulated as 
\begin{equation}
\bm{\Lambda}_d \ddot{\tilde{\bm{{x}}}}+\bm{D}_d \dot{\tilde{\bm{{x}}}}+\bm{K}_d \tilde{\bm{{x}}} =-\bm{K}_f \tilde{\bm{f}}+
\bm{\Lambda}_d (\dot{\bm{s}}+\bm{A}\bm{s})
\end{equation}
where in the sliding mode (i.e., $\bm{s}\rightarrow0,~\dot{\bm{s}}\rightarrow0$), the desired dynamics of end-effector \eqref{desired dynamics} is achieved.

As such, the control law $\bm{F}_{\tau}$ is synthesized to reach the sliding mode. The latter includes a model compensation term $\bm{F}_{\tau m}$, a linear stabilize feedback term $\bm{F}_{\tau s1}$ and z robust feedback term $\bm{F_{\tau s2}}$, specifically as 
\begin{equation} \label{input}
    \begin{aligned}
        &\bm{F}_{\tau} = \bm{F}_{\tau m} + \bm{F}_{\tau s1}+ \bm{F}_{\tau s2}\\
        &\bm{F}_{\tau m} = \bm{\Lambda}(\bm{x})\ddot{\bm{x}}_{eq}+\bm{\mu}(\bm{x},\dot{\bm{x}})\dot{\bm{x}}_{eq}+\bm{F}_{g}+\hat{\bm{D}} -\bm{F}_{ext}\\
        &\bm{F}_{\tau s1} = -\bm{K}_{\bm{s}}\bm{s}\\
        &\bm{F}_{\tau s2} = -\frac{1}{4\eta }h^2 \bm{s}
    \end{aligned}
\end{equation}
where $\dot{\bm{x}}_{eq}=\dot{\bm{x}}_d - \bm{F}_1 \tilde{\bm{x}}-\bm{F}_2\bm{z}$ and $\ddot{\bm{x}}_{eq}=\ddot{\bm{x}}_d - \bm{F}_1 \dot{\tilde{\bm{x}}}-\bm{F}_2\dot{\bm{z}}$; $\hat{\bm{D}}$ is the estimated value of lumped uncertainties which is designed below; $\bm{K}_{\bm{s}}$ is the positive definite matrix that stabilizing the system; $\eta$ and $h$ are positive scalars with $h$ satisfying $h\geq \|\bm{D}-\hat{\bm{D}}\|$.

To obtain the estimated value of lumped uncertainties $\hat{\bm{D}}$, the adaptation law is designed as following,
\begin{equation} \label{D}
  \dot{\hat{\bm{D}}} = - {Proj}_{\hat{{D}}}(\bm{\gamma}_d \bm{s}),\ |\hat{\bm{D}}(0)|\leq \hat{\bm{D}}_{max}
\end{equation}
where $\bm{\gamma}_d$ is a diagonal s.p.d. matrix and $\hat{\bm{D}}_{max}$ is the bound for $\hat{\bm{D}}$. The projection-type mapping in \eqref{D} guarantees that $|\hat{\bm{D}}(t)|\leq \hat{\bm{D}}_{max},\ \forall t$.
\begin{figure*}[thpb]
    \centering
    \subfigure[]{\includegraphics[width=0.49\textwidth]{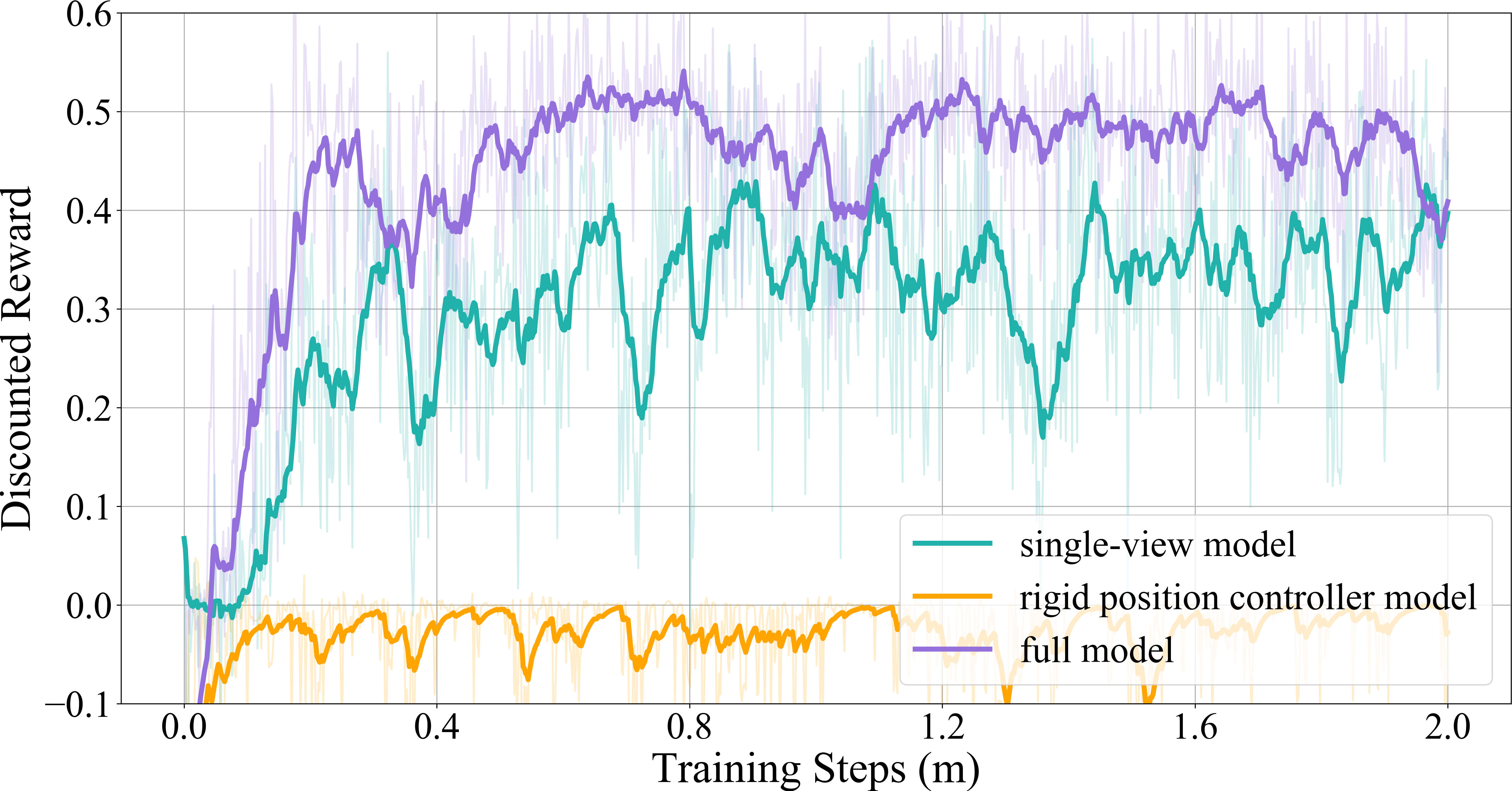}} 
    \subfigure[]{\includegraphics[width=0.49\textwidth]{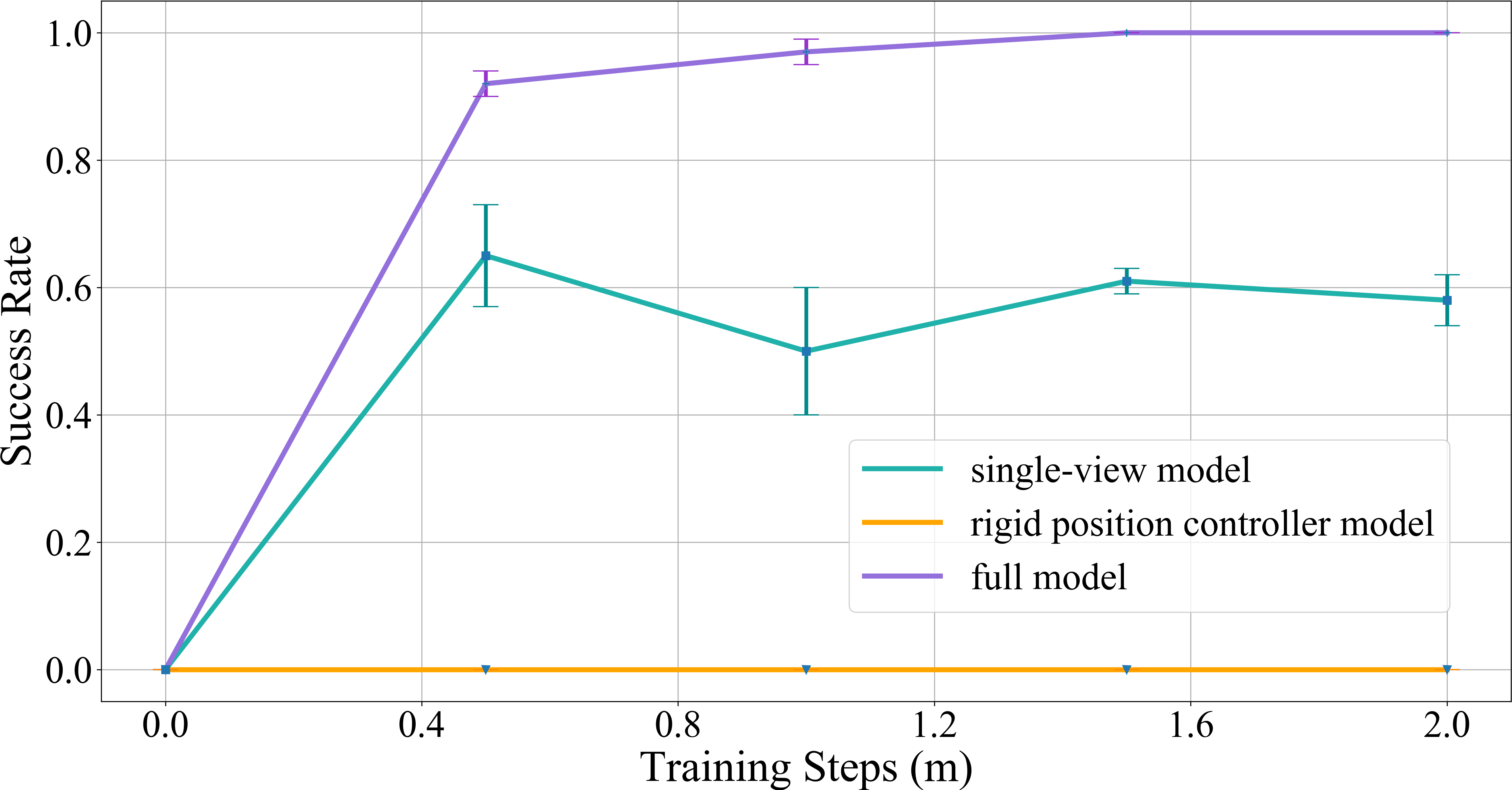}} \\

    \caption{Comparative studies of the procession of the multi-modality. (a) Training curves of three models, including the single-view model, the rigid position controller model, and the full model. (b) The insertion success rates at different training stages of three models.
    }
    \label{ablative}
\end{figure*}
The above control law \eqref{input} is synthesized in operational space, while joint space control input $\bm{\tau}$ is required in practice. Therefore, extract the transpose pseudo inverse of Jacobian matrix $\bm{J}^{\dagger T}$ from \eqref{input}, we arrive at 
\begin{equation}
    \begin{aligned}
        \bm{J}^{\dagger T}\bm{\tau} = \bm{F}_{\tau} = \bm{J}^{\dagger T}(&{\bm{M}}(\bm{q})\ddot{\bm{q}}_{eq}+{\bm{C}}(\bm{q},\dot{\bm{q}})\dot{\bm{q}}_{eq}+{\bm{G}}(\bm{q}) \\
        &+\bm{J}^T(\hat{\bm{D}}-\bm{F}_{ext}-\bm{K}_s \bm{s}-\frac{1}{4\eta }h^2 \bm{s}))
    \end{aligned}
\end{equation}
where $\dot{\bm{q}}_{eq}=\bm{J}^{\dagger}\dot{\bm{x}}_{eq}$ and $\ddot{\bm{q}}_{eq}=\bm{J}^{\dagger}(\ddot{\bm{x}}_{eq}-\dot{\bm{J}}\dot{\bm{q}}_{eq})$.

\subsection{Hierarchical Policy Learning}
\label{subsec:par2}
\label{subsec:hierarchial learning}
The multi-modal observation is high-dimensional and challenging for the RL algorithm to learn useful policies. Therefore, a hierarchical policy learning approach is proposed.

The hierarchical policy learning approach can be divided into two stages. During the first stage, a visuomotor policy is developed with the extracted visual features. The action space is three dimensional representing the desired residual displacement along $\bm{x}$ and $\bm{y}$ directions and the rotation about the $\bm{z}$-axis (\figref{detail_scene}). We adopt an object-centric description of the action space to make the control scheme unaffected by the configuration of robot. 
As shown in \eqref{action1}, the desired position of the gripper is deducted by adding the command position to the current position. 
\begin{equation}\label{action1}
    \setlength{\abovedisplayskip}{3pt}
    \setlength{\belowdisplayskip}{3pt}
    \bm{pos_d} = \bm{pos_c} + \bm{rot} * \Delta \bm{x}
\end{equation}
Where $\bm{pos_d}$ and $\bm{pos_c}$ denote the desired and the current positions of the gripper,  $\bm{rot}$ characterizes the transformation matrix of the gripper, and $\Delta \bm{x}$ is the position command in the object frame.
The desired orientation of the gripper is deducted by adding the command rotation to the current position as shown in \eqref{action2}. 
\begin{equation}\label{action2}
    \setlength{\abovedisplayskip}{3pt}
    \setlength{\belowdisplayskip}{3pt}
    \bm{rot_d} = \bm{rot} * \begin{bmatrix} \bm{\cos{\bm{\theta}}}& -\bm{\sin{\bm{\theta}}} &0 \\ \bm{\sin{\bm{\theta}}} & \bm{\cos{\bm{\theta}}} & 0 \\ 0 & 0& 1
    \end{bmatrix}
\end{equation}
Where $\bm{rot_d}$ denotes the desired orientation of the gripper in the hybrid motion/force controller. $\bm{\theta}$ is the command rotation signal.

The second stage of training is built on the result of the first stage. Specifically, the multi-modal policy parameters in the second stage are initialized with the visuomotor policy (result of the first stage).
The observation space is augmented with the fusion of the 6-axis force and the 6-dimensional velocity (\figref{policy}). Simultaneously, the desired displacement along the $\bm{z}$-axis and the stiffness along the insertion direction are actively modified. The two-stage training procedure is detailed in \figref{policy}.

\section{EXPERIMENTS: DESIGN AND SETUP}
\label{sec:section5}
The primary goal of our experiments is to examine the efficacy of the hierarchically learned multi-modal control scheme in simulated and real-world contact-rich tasks. In particular, we design the experiments to answer the following research questions:
\begin{itemize}
    \item RQ1. How does the present work behave compared to the state-of-the-art multi-modal manipulation policies?
    \item RQ2. What are the effects of the multi modalities?
    \item RQ3. What is the advantage of the hierarchical policy learning?
    \item RQ4. Can the proposed method be generalized to unseen shapes and real scenarios?
\end{itemize}

\subsection{Task Description}
We design a series of assembly tasks as shown in \figref{generalization} and \figref{fig9}. In the simulation, seven shapes are used, including square, circular, pentagonal, triangular, and other shapes. The clearance of the pegs and the holes can be narrowed to 0.25mm in radius.
During training, the position and orientation of the peg and hole are sampled from uniform distributions. The initial position offsets of the peg and hole along $\bm{x}$, $\bm{y}$ and $\bm{z}$ axis could be up to 4cm, 4cm, and 5cm. The misalignment of the rotation of the peg and the hole can be up to 60 degrees.
Meanwhile, when tested in a simulated or real environment, the initial position and rotation offsets are consistent with the training conditions.

\subsection{Robot Environment Setup}
For both simulation and real robot experiments, we use the Franka Emika Panda robot. Three information modalities are utilized, including vision, force, and proprioceptive data. 
In the simulation, force data are acquired from the F/T sensor mounted on the wrist of the robot. In addition, multi-view data are provided by two cameras mounted on the top of the end-effector. The simulated setting is illustrated in \figref{fig1}.
Similarly, two Intel Realsense D435 cameras are mounted on the wrist of the robot in real experiments.
The ATI F/T sensor is installed between the camera and the gripper. Moreover, soft fingertips made by Dragon Skin are placed on the gripper to better fit the shapes of objects. The real scene setting is illustrated in \figref{fig9}.
\begin{figure}[thpb]
    \centering
    \includegraphics[scale=0.27]{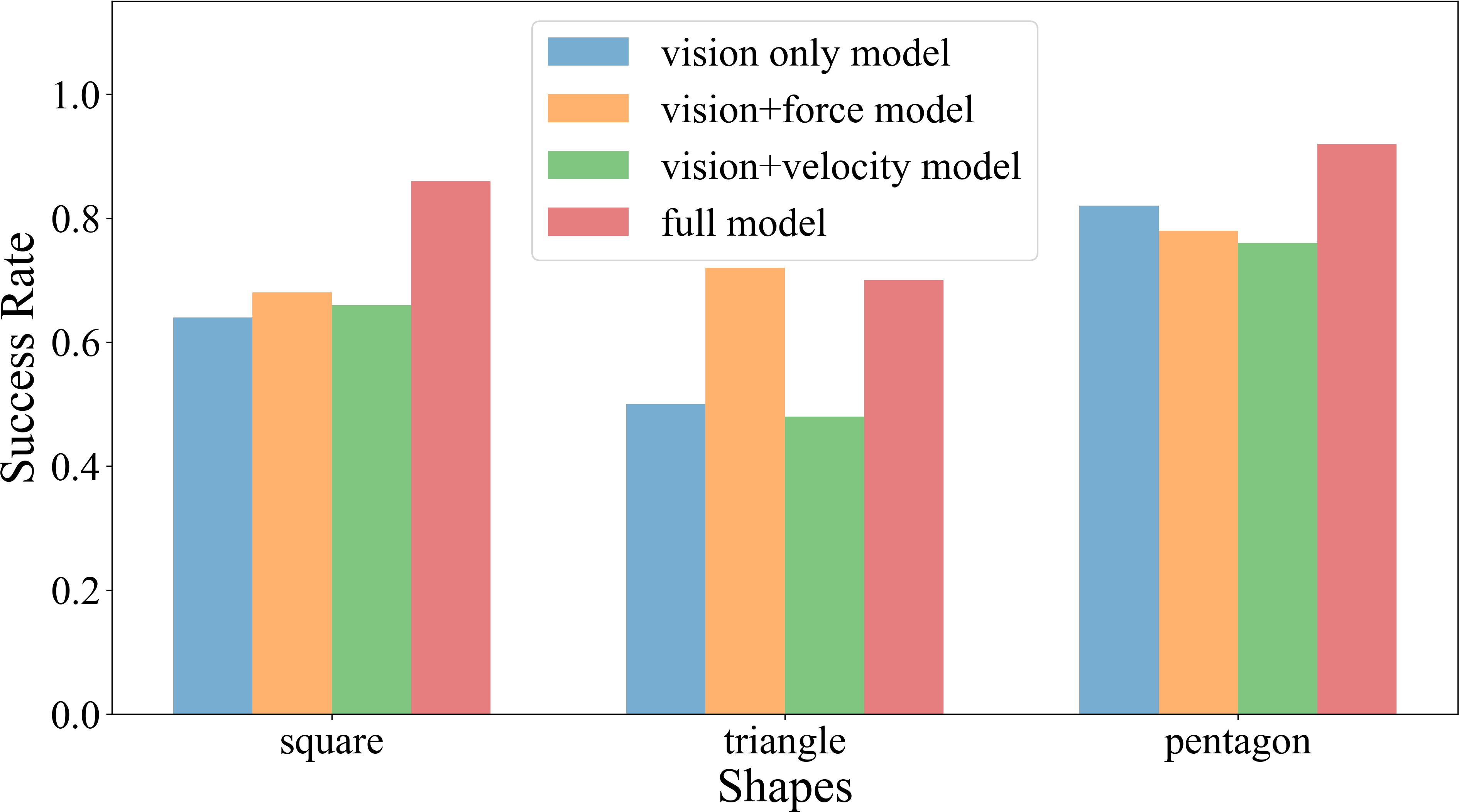}
    \caption{Ablative study of multi-modality fusions in the second training stage. The full model shows better performance on trained shape (square) and unseen shapes (triangle and pentagon).
    }
    \label{second_ablative}
\end{figure}
\subsection{Self-supervised Image Encoder}
To collect the dataset for the self-supervised visual encoder, a stochastic policy is run in simulation. The collected 60k pictures are first cropped into the size of 224 $\times$ 224 as shown in \figref{fig4}.
The encoder is trained with a batch size of 64 and a learning rate of 0.001. 

\subsection{ARMRC-based Hybrid Motion/Force Control Setup}
The hybrid motion/force controller is motion-controlled along the $\bm{x}$-$\bm{y}$ plane while force controlled along the $\bm{z}$-axis.
The initial setting is achieved by choosing the desired dynamics as in \eqref{desired dynamics} with the desired inertia matrix $\bm{\Lambda}_d = diag[1.0,1.0,1.0,0.01,0.01,0.01]$, the desired stiffness matrix $\bm{K}_d = diag[500,500,100,1.0,1.0,1.0]$ the desired damping matrix $ \bm{D_d}=\bm{2\sqrt{K_d M_d^{-1}}M_d}$
and the desired force gain matrix $\bm{K_f} = diag[0.0,0.0,-0.5,0,0,0]$.
The desired force along $\bm{z}$-axis is $\bm{f_d}=0.5N$.

The hybrid motion/force controller parameters are modified during the second training stage.
Specifically, the stiffness along the $\bm{z}$-axis is learned through RL. The desired stiffness matrix is rewritten as $\bm{K}_d = diag[500,500,100+\alpha * (500-100),1.0,1.0,1.0]$. $\alpha$ is one of the action outputs of the policy. The desired damping matrix is rewritten accordingly.

\subsection{Hierarchical Policy Learning Setup}
In the hierarchical policy learning, the rewards at two stages are designed differently.
In the first stage, the agent gets the reward $\bm{1}$ if the task is finished successfully, penalty $\bm{-1}$ if the object is off the hole plane. 
In the second stage, the agent gets the reward $\bm{0.5}$ if the object is half in the hole and another $\bm{1}$ if the task is finished successfully. The agent will get a penalty $\bm{-0.1}$ if the object was once in the half of the hole but fails the task. 
The RL algorithm is trained using Proximal Policy Optimization (PPO) \cite{schulman2017proximal}, implemented by Stable baselines \cite{hill2018stable}.
The policy neural network has two hidden layers, with each having 32 neurons.
While training, the whole task has a time constraint of 200-time steps, each lasts 0.02 s. 

\begin{figure}[thpb]
    \centering
    \includegraphics[scale=0.27]{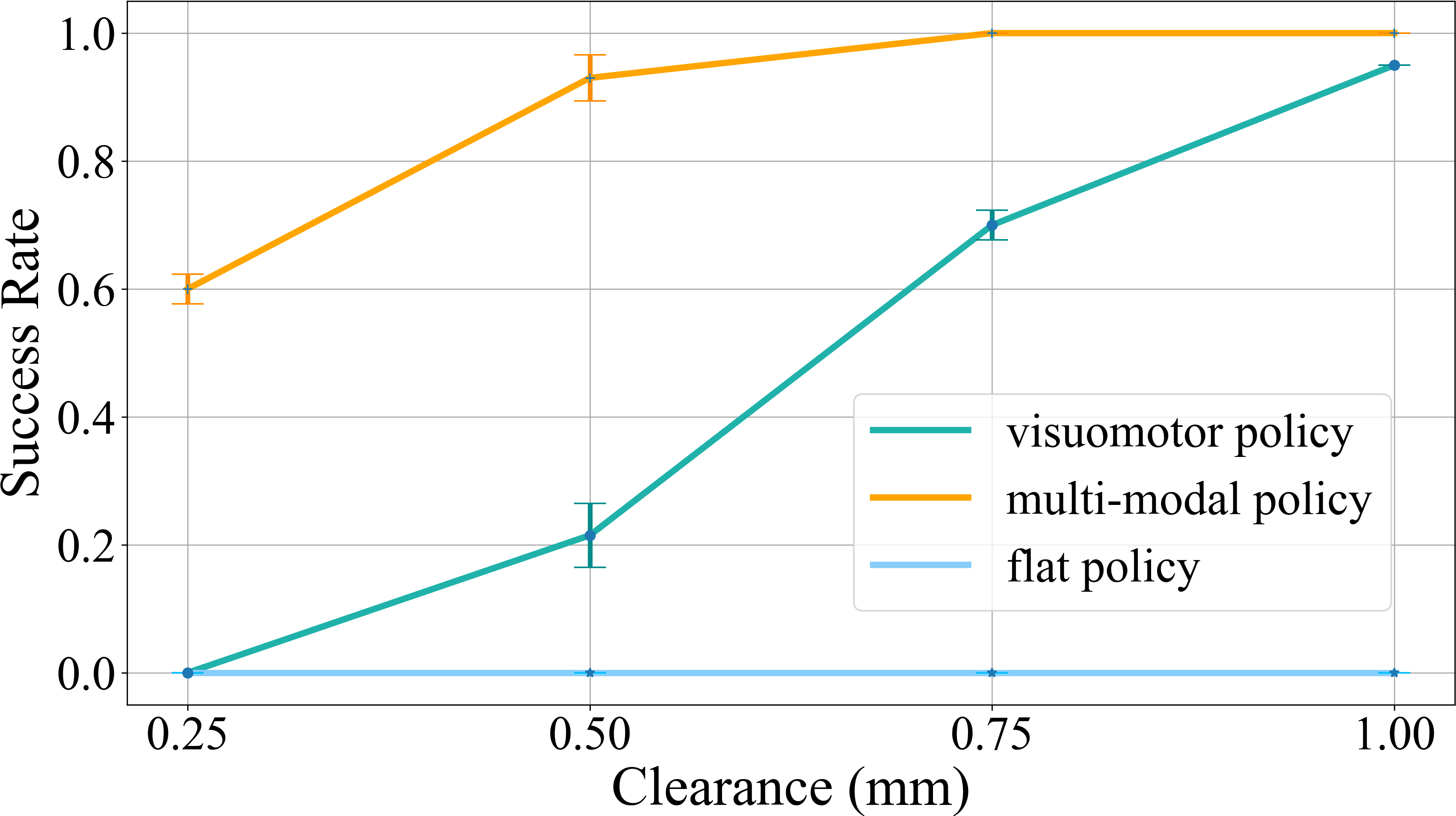}
    \caption{Comparisons of the hierarchical policy learning approach and flat policy learning approach on four different clearances.
    The hierarchical policy is comprised of two stages: the visuomotor policy and the multi-modal policy. 
    }
    \label{one_two_comp}
\end{figure}

\section{EXPERIMENTS: RESULTS}
This section first conducts comparisons between the proposed method with the current state-of-the-art model\cite{lee2020making}. 
Next, the contributions of multi-modalities are studied. Furthermore, the hierarchical policy learning approach is compared with the flat learning approach. Finally, the generalization ability to the new shapes and real scenarios are explored.
\subsection{Performance Comparisons (RQ1)}
Few works studying the fusion of vision, force and proprioception in the assembly task. Therefore, we only experimentally compare our method with the multi-modal controller in \cite{lee2020making} (shown in Table \ref{sota}). From the table one could see that the present method achieves better results with less modality input (no depth feedback).
\begin{table}[h]
	\caption{The performance of different multi-modal models in the assembly task}
	\label{Table: Ablation exp}
	\begin{center}
		\begin{tabular}{|c|c|c|c|c|c|}
		    \hline
			Models & Clearance & \makecell[c]{Success \\ Rate} & Depth &  \makecell[c]{Initial \\Orientation \\ Mismatch}   & Sim2real\\
			\hline
			Lee \cite{lee2020making}  & 2.24mm  & 78$\%$  & Yes & No & No\\
			\hline
			Ours & 0.5mm & 92$\%$ & No & Yes & Yes\\
			\hline
			
		\end{tabular}
	\end{center}
	\label{sota}
\end{table}

\subsection{Multi-modality Ablations (RQ2)}
In this experiment, we mainly exploit the effects of the multi-modality in two designs, namely, the multi-modality procession and the multi-modal representation fusion.
The square assembly task is used for evaluation. 
\label{subsec:multi-modality procession}
\subsubsection{Multi-modality Procession}
\begin{itemize}
    \item \textbf{Single-view model} only utilizes the front camera in the self-supervised visual encoder.
    \item \textbf{Rigid position controller model} uses the motion control part of the hybrid controller and ignore the force control component.
    \item \textbf{Full model} contains a multi-view self-supervised visual encoder and a hybrid motion/force controller.
\end{itemize}
\begin{figure}[thpb]
    \centering
    \includegraphics[scale=0.58]{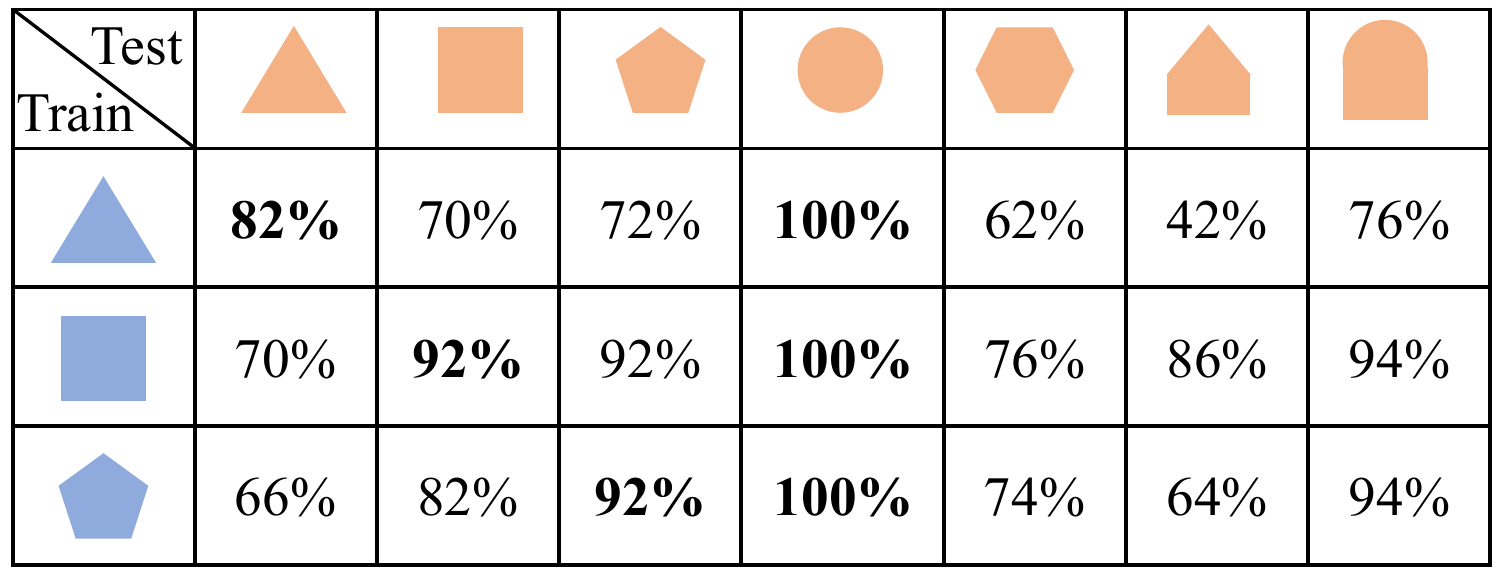}
    \caption{
    Generalization performance of the multi-modal control scheme. Three shapes (\textcolor{train_shape}{blue}) are used in the training process, the triangular, square, and pentagonal peg. The generalization ability of control scheme to seven novel shapes (\textcolor{test_shape}{orange}) is tested. 
    }
    \label{generalization}
\end{figure}

The training curves and the test results are illustrated in \figref{ablative}.
The test results are acquired with five random seeds, each with 20 trials.
The result demonstrates that the performance of the full model substantially exceeds the performance of the compared models with single-view or motion controller alone. 
The multi-view setup contains symmetric images and more information, thereby providing more robust spatial features.
Meanwhile, the force controller not only eases the danger of collision between the robot and the environment but also reduces the surface friction, which resists the planar movement of the robot.
\subsubsection{Multi-modal Representation Fusion}
\label{subsec:ablative}
\begin{itemize}
    \item \textbf{Vision only model} contains only visual feedback in the observation space.
    \item \textbf{Vision+force model} fuses the visual and force feedback.
    \item \textbf{Vision+velocity model} fuses the visual feedback and velocity.
    \item \textbf{Full model} fuses the visual, force and velocity feedback.
\end{itemize}

Two tasks are performed. The control scheme is trained with a square object. To evaluate the performance of the model, we respectively test the control scheme on the square and the other two unseen shapes, the triangular and the pentagonal shape. 
The test results shown in \figref{second_ablative} are acquired with 50 trials.
We analyze that the full model shows better and more robust behavior on seen and unseen shapes. The redundant information incorporation facilitates the execution under more extreme conditions.

\subsection{Policy Learning Structure (RQ3)}
In this experiment, we mainly analyze the advantages of the hierarchical policy learning approach. 
Firstly, the performance improvement from visuomotor policy to multi-modal policy is studied. Secondly, the hierarchical policy learning approach is compared with the flat policy learning approach. We compare the performance of various models in a square assembly task with four clearances, i.e., 0.25mm, 0.5mm, 0.75mm, and 1mm. 
\label{subsec:hierarchical learning}

\subsubsection{Visuomotor Policy vs. Multi-modal Policy}
In the hierarchical policy learning method, the first-stage visuomotor policy is further improved by force and proprioceptive feedback incorporation. In addition, the stiffness along the insertion direction is actively learned in the second stage.
The comparison results can be seen in \figref{one_two_comp}. The test results are acquired with five random seeds, each with 20 trials.
From the result, we could see that the performance is further enhanced in the second stage. 
\subsubsection{Hierarchical Policy Learning vs. Flat Policy Learning}
The hierarchical learning method is compared with the flat policy learning. The latter has the same structure as the one in the current policy learning method. Rather than learning policy hierarchically, the flat approach tries to learn the policy in one single stage.
The test result can be seen in \figref{one_two_comp}. We observe that the reinforcement learning (RL) algorithm fails to learn directly the manipulation policies from the complex observation and action space. The hierarchical policy learning approach is thus essential in multi-modal manipulation policy learning.
\subsection{Generalization Test (RQ4)}
\label{subsec:generalization}
In this experiment, we examine the potential of transferring the learned policies to new shapes. Seven shapes triangular, square, pentagonal, circular, and other pegs shown in \figref{generalization} are used with 0.5mm tolerance. We transfer the control scheme learned with one shape to the other six shapes in the simulation.
In each test, 50 trials are conducted. From \figref{generalization}, one finds that the multi-modal control scheme could extract mutual information enabling the gripper to move and rotate to accomplish the assembly task, regardless the differences in shapes and sizes.
Most fails are due to the abrupt force behavior related to the contact model in the simulation. The relatively poor behavior for the triangular shape might attribute to its sharp edges.


\begin{figure}[thpb]
    \centering
    \includegraphics[scale=0.5]{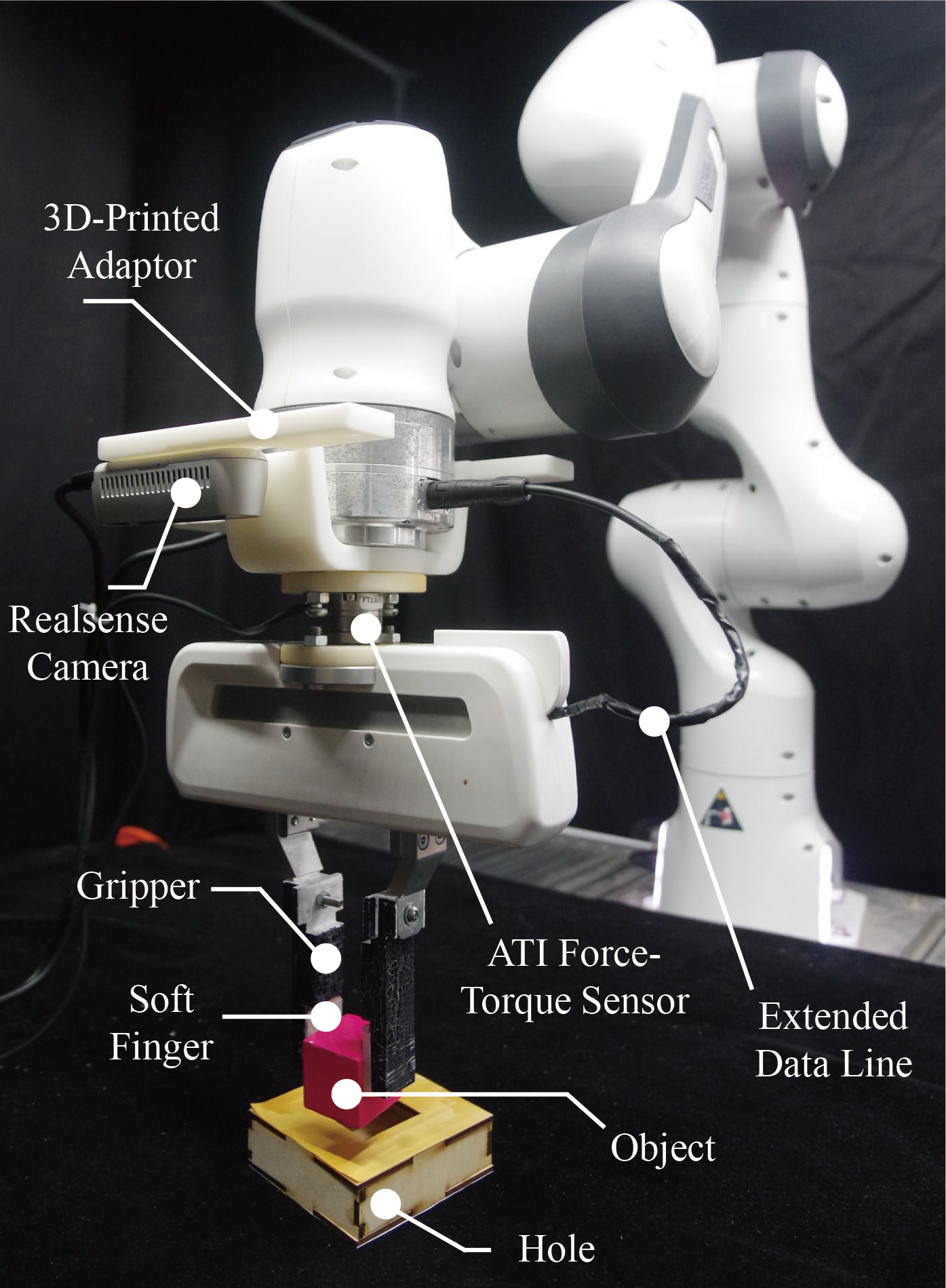}
    \caption{The real scene of the task.
    }
    \label{fig9}
\end{figure}
\subsection{Real Robot Experiments (RQ4)}
\label{subsec:real}
The uncertainties of the sim-to-real transfer mainly come from two sources, the sensor noise and the control uncertainty.
Due to the varied lights conditions and objects textures in the simulation and reality, the vision encoder learned in simulation might be invalid in real situations. Similarly, the force feedback has different characteristics in simulated and the real conditions. Moreover, the model inconsistency from the simulation to the real scenario introduces extra challenges to the high-frequency motion/force hybrid controller.
To deploy directly the control scheme on a real robot, domain randomization is applied in simulation. We randomize the textures and colors of the grasped object, hole plane, and backgrounds. Besides, the hybrid motion/force controller is designed to be robust with model uncertainty. Two main tests are conducted, including the direct sim-to-real transfer and the robustness test. The results can be seen in the supplementary videos.
\subsubsection{Direct sim-to-real transfer}
Without further tuning the parameters, the control scheme is transferred to the real environments with unknown target positions and large position and orientation mismatches.

\subsubsection{Robustness test}
To further evaluate the robustness of the control scheme, the hole is randomly placed and moved while testing. Besides, the images are randomly blocked while the robot executes the task. 

\section{DISCUSSION AND CONCLUSION}
In this work, a hierarchical policy learning approach for multi-modal fusion is validated on the precise multi-shape assembly task.
The hierarchical utilization of the vision, force and proprioceptive data simplifies the construction of the modality encoders and improves the resultant control scheme.
The control scheme generalizes to unseen configurations, new shapes, and real-world scenarios.

In developing the control scheme, we find that the effectiveness of the control scheme is highly related to the parameters in the hybrid motion/force controller. Perception and control are highly coupled in contact-rich tasks. As much we focus on the sensor fusion, a stable and qualified low-level controller is decisive for task completion.

In future work, we plan to deploy the control scheme on unknown task space.
\section{Acknowledgment}
This work acknowledges the supports by the following programs: National Key R \& D Program of China 2017 YFA0701100, 
National Natural Science Foundation of China (T2125009, 11822207,  92048302), 
Zhejiang Provincial Natural Science Foundation of China (R18A020004).

\bibliographystyle{IEEEtran}
\bibliography{paper}

@article{andrychowicz2020learning,
  title={Learning dexterous in-hand manipulation},
  author={Andrychowicz, OpenAI: Marcin and Baker, Bowen and Chociej, Maciek and Jozefowicz, Rafal and McGrew, Bob and Pachocki, Jakub and Petron, Arthur and Plappert, Matthias and Powell, Glenn and Ray, Alex and others},
  journal={The International Journal of Robotics Research},
  volume={39},
  number={1},
  pages={3--20},
  year={2020},
  publisher={SAGE Publications Sage UK: London, England}
}

@inproceedings{van2016stable,
  title={Stable reinforcement learning with autoencoders for tactile and visual data},
  author={Van Hoof, Herke and Chen, Nutan and Karl, Maximilian and van der Smagt, Patrick and Peters, Jan},
  booktitle={2016 IEEE/RSJ international conference on intelligent robots and systems (IROS)},
  pages={3928--3934},
  year={2016},
  organization={IEEE}
}

@inproceedings{luo2019reinforcement,
  title={Reinforcement learning on variable impedance controller for high-precision robotic assembly},
  author={Luo, Jianlan and Solowjow, Eugen and Wen, Chengtao and Ojea, Juan Aparicio and Agogino, Alice M and Tamar, Aviv and Abbeel, Pieter},
  booktitle={2019 International Conference on Robotics and Automation (ICRA)},
  pages={3080--3087},
  year={2019},
  organization={IEEE}
}

@article{schulman2017proximal,
  title={Proximal policy optimization algorithms},
  author={Schulman, John and Wolski, Filip and Dhariwal, Prafulla and Radford, Alec and Klimov, Oleg},
  journal={arXiv preprint arXiv:1707.06347},
  year={2017}
}

@book{sutton2018reinforcement,
  title={Reinforcement learning: An introduction},
  author={Sutton, Richard S and Barto, Andrew G},
  year={2018},
  publisher={MIT press}
}

@article{hill2018stable,
  title={Stable baselines},
  author={Hill, Ashley and Raffin, Antonin and Ernestus, Maximilian and Gleave, Adam and Traore, Rene and Dhariwal, Prafulla and Hesse, Christopher and Klimov, Oleg and Nichol, Alex and Plappert, Matthias and others},
  journal={GitHub repository},
  year={2018},
  publisher={GitHub}
}

@article{Calandra_2018,
title={More Than a Feeling: Learning to Grasp and Regrasp Using Vision and Touch},
volume={3},
ISSN={2377-3774},
url={http://dx.doi.org/10.1109/LRA.2018.2852779},
DOI={10.1109/lra.2018.2852779},
number={4},
journal={IEEE Robotics and Automation Letters},
publisher={Institute of Electrical and Electronics Engineers (IEEE)},
author={Calandra, Roberto and Owens, Andrew and Jayaraman, Dinesh and Lin, Justin and Yuan, Wenzhen and Malik, Jitendra and Adelson, Edward H. and Levine, Sergey},
year={2018},
month={Oct},
pages={3300–3307}
}

@article{10.1115/1.3149634,
    author = {Whitney, D. E.},
    title = "{Quasi-Static Assembly of Compliantly Supported Rigid Parts}",
    journal = {Journal of Dynamic Systems, Measurement, and Control},
    volume = {104},
    number = {1},
    pages = {65-77},
    year = {1982},
    month = {03},
    issn = {0022-0434},
    doi = {10.1115/1.3149634},
    url = {https://doi.org/10.1115/1.3149634},
    eprint = {https://asmedigitalcollection.asme.org/dynamicsystems/article-pdf/104/1/65/5777806/65\_1.pdf},
}

@article{DBLP:journals/corr/abs-2102-06279,
  author    = {Wei Gao and
               Russ Tedrake},
  title     = {kPAM 2.0: Feedback Control for Category-Level Robotic Manipulation},
  journal   = {CoRR},
  volume    = {abs/2102.06279},
  year      = {2021},
  url       = {https://arxiv.org/abs/2102.06279},
  archivePrefix = {arXiv},
  eprint    = {2102.06279},
  bibsource = {dblp computer science bibliography, https://dblp.org}
}

@article{DBLP:journals/corr/abs-1803-09956,
  author    = {Andy Zeng and
               Shuran Song and
               Stefan Welker and
               Johnny Lee and
               Alberto Rodriguez and
               Thomas A. Funkhouser},
  title     = {Learning Synergies between Pushing and Grasping with Self-supervised
               Deep Reinforcement Learning},
  journal   = {CoRR},
  volume    = {abs/1803.09956},
  year      = {2018},
  url       = {http://arxiv.org/abs/1803.09956},
  archivePrefix = {arXiv},
  eprint    = {1803.09956},
  bibsource = {dblp computer science bibliography, https://dblp.org}
}

@article{luo2021robust,
  title={Robust Multi-Modal Policies for Industrial Assembly via Reinforcement Learning and Demonstrations: A Large-Scale Study},
  author={Luo, Jianlan and Sushkov, Oleg and Pevceviciute, Rugile and Lian, Wenzhao and Su, Chang and Vecerik, Mel and Ye, Ning and Schaal, Stefan and Scholz, Jon},
  journal={arXiv preprint arXiv:2103.11512},
  year={2021}
}

@inproceedings{dong2021icra,

 title={Tactile-RL for Insertion: Generalization to Objects of Unknown Geometry},

 author={Dong, Siyuan and Jha, Devesh and Romeres, Diego and Kim, Sangwoon and Nikovski, Daniel and Rodriguez, Alberto},

  booktitle={2021 IEEE International Conference on Robotics and Automation (ICRA)},

  year={2021},

  URL={https://arxiv.org/pdf/2104.01167.pdf} 

}

@INPROCEEDINGS{7743375,  author={Tang, Te and Lin, Hsien-Chung and Yu Zhao and Wenjie Chen and Tomizuka, Masayoshi},  booktitle={2016 IEEE International Conference on Automation Science and Engineering (CASE)},   title={Autonomous alignment of peg and hole by force/torque measurement for robotic assembly},   year={2016},  volume={},  number={},  pages={162-167},  doi={10.1109/COASE.2016.7743375}}

@ARTICLE{9361338,  author={Oikawa, Masahide and Kusakabe, Tsukasa and Kutsuzawa, Kyo and Sakaino, Sho and Tsuji, Toshiaki},  journal={IEEE Robotics and Automation Letters},   title={Reinforcement Learning for Robotic Assembly Using Non-Diagonal Stiffness Matrix},   year={2021},  volume={6},  number={2},  pages={2737-2744},  doi={10.1109/LRA.2021.3060389}}

@inproceedings{lee2020multimodal,
  title={Multimodal sensor fusion with differentiable filters},
  author={Lee, Michelle A and Yi, Brent and Mart{\'\i}n-Mart{\'\i}n, Roberto and Savarese, Silvio and Bohg, Jeannette},
  booktitle={2020 IEEE/RSJ International Conference on Intelligent Robots and Systems (IROS)},
  pages={10444--10451},
  year={2020},
  organization={IEEE}
}

@article{narita2021policy,
  title={Policy Blending and Recombination for Multimodal Contact-Rich Tasks},
  author={Narita, Tetsuya and Kroemer, Oliver},
  journal={IEEE Robotics and Automation Letters},
  volume={6},
  number={2},
  pages={2721--2728},
  year={2021},
  publisher={IEEE}
}

@article{DBLP:journals/corr/abs-1809-08564,
  author    = {Douglas Morrison and
               Peter Corke and
               J{\"{u}}rgen Leitner},
  title     = {Multi-View Picking: Next-best-view Reaching for Improved Grasping
               in Clutter},
  journal   = {CoRR},
  volume    = {abs/1809.08564},
  year      = {2018},
  url       = {http://arxiv.org/abs/1809.08564},
  eprinttype = {arXiv},
  eprint    = {1809.08564},
  bibsource = {dblp computer science bibliography, https://dblp.org}
}

@inproceedings{zhang2020visual,
  title={Visual tactile fusion object clustering},
  author={Zhang, Tao and Cong, Yang and Sun, Gan and Wang, Qianqian and Ding, Zhenming},
  booktitle={Proceedings of the AAAI Conference on Artificial Intelligence},
  volume={34},
  number={06},
  pages={10426--10433},
  year={2020}
}

@INPROCEEDINGS{9561222,  author={von Drigalski, Felix and Hayashi, Kennosuke and Huang, Yifei and Yonetani, Ryo and Hamaya, Masashi and Tanaka, Kazutoshi and Ijiri, Yoshihisa},  booktitle={2021 IEEE International Conference on Robotics and Automation (ICRA)},   title={Precise Multi-Modal In-Hand Pose Estimation using Low-Precision Sensors for Robotic Assembly},   year={2021},  volume={},  number={},  pages={968-974},  doi={10.1109/ICRA48506.2021.9561222}}

@inproceedings{li2019connecting,
  title={Connecting touch and vision via cross-modal prediction},
  author={Li, Yunzhu and Zhu, Jun-Yan and Tedrake, Russ and Torralba, Antonio},
  booktitle={Proceedings of the IEEE/CVF Conference on Computer Vision and Pattern Recognition},
  pages={10609--10618},
  year={2019}
}

@inproceedings{lee2021detect,
  title={Detect, reject, correct: Crossmodal compensation of corrupted sensors},
  author={Lee, Michelle A and Tan, Matthew and Zhu, Yuke and Bohg, Jeannette},
  booktitle={2021 IEEE International Conference on Robotics and Automation (ICRA)},
  pages={909--916},
  year={2021},
  organization={IEEE}
}

@article{lecun1995convolutional,
  title={Convolutional networks for images, speech, and time series},
  author={LeCun, Yann and Bengio, Yoshua and others},
  journal={The handbook of brain theory and neural networks},
  volume={3361},
  number={10},
  pages={1995},
  year={1995}
}

@inproceedings{vaswani2017attention,
  title={Attention is all you need},
  author={Vaswani, Ashish and Shazeer, Noam and Parmar, Niki and Uszkoreit, Jakob and Jones, Llion and Gomez, Aidan N and Kaiser, {\L}ukasz and Polosukhin, Illia},
  booktitle={Advances in neural information processing systems},
  pages={5998--6008},
  year={2017}
}

@article{lee2020making,
  title={Making sense of vision and touch: Learning multimodal representations for contact-rich tasks},
  author={Lee, Michelle A and Zhu, Yuke and Zachares, Peter and Tan, Matthew and Srinivasan, Krishnan and Savarese, Silvio and Fei-Fei, Li and Garg, Animesh and Bohg, Jeannette},
  journal={IEEE Transactions on Robotics},
  volume={36},
  number={3},
  pages={582--596},
  year={2020},
  publisher={IEEE}
}

\end{document}